# Enhancing personalised thermal comfort models with Active Learning for improved HVAC controls

**Zeynep Duygu Tekler, Yue Lei, Xilei Dai, Adrian Chong**

Department of the Built Environment, National University of Singapore, Singapore

E-mail: `zdtekler@nus.edu.sg`

**Abstract.** Developing personalised thermal comfort models to inform occupant-centric controls (OCC) in buildings requires collecting large amounts of real-time occupant preference data. This process can be highly intrusive and labour-intensive for large-scale implementations, limiting the practicality of real-world OCC implementations. To address this issue, this study proposes a thermal preference-based HVAC control framework enhanced with Active Learning (AL) to address the data challenges related to real-world implementations of such OCC systems. The proposed AL approach proactively identifies the most informative thermal conditions for human annotation and iteratively updates a supervised thermal comfort model. The resulting model is subsequently used to predict the occupants' thermal preferences under different thermal conditions, which are integrated into the building's HVAC controls. The feasibility of our proposed AL-enabled OCC was demonstrated in an EnergyPlus simulation of a real-world testbed supplemented with the thermal preference data of 58 study occupants. The preliminary results indicated a significant reduction in overall labelling effort (i.e., 31.0%) between our AL-enabled OCC and conventional OCC while still achieving a slight increase in energy savings (i.e., 1.3%) and thermal satisfaction levels above 98%. This result demonstrates the potential for deploying such systems in future real-world implementations, enabling personalised comfort and energy-efficient building operations.

## 1. Introduction
Personal thermal comfort models are used to predict individual-level thermal comfort responses and can capture the distinct preferences between individuals compared to aggregated comfort models [1]. Integrating these models into HVAC controls is essential for occupant-centric controls (OCC) and has been shown in past studies to significantly reduce HVAC energy consumption while enhancing occupants' thermal satisfaction through various modelling approaches and control strategies. For example, Jazizadeh et al. [2] developed a fuzzy-rule-based model to determine the preferred temperature of each occupant and adjusted the centralised HVAC temperature setpoint via BMS to minimize the sum differences between local temperatures and preferred temperatures in each thermal zone. As a result, the daily airflow rates were reduced by 39%, compared to conventional HVAC controls with predefined temperature setpoints, while significantly improving thermal satisfaction. A study conducted by Aryal et al. [3] proposed the use of Naïve Bayes-based probabilistic thermal comfort models to inform the HVAC temperature setpoint, resulting in a 25% average increase in occupant satisfaction and a 2.1% increase in energy savings compared to a fixed temperature setpoint of 22.5°C.

Despite the benefits of integrating personal thermal comfort models with HVAC systems to achieve OCC, the real-world practicality of such implementations remains limited as these personalised comfort models require large amounts of occupants' preference data to achieve accurate model performances. This preference data is usually collected using various surveying tools, such as online surveys and wearables, which are often highly intrusive and labour-intensive, especially when conducting large-scale data collection studies [1]. The challenge related to data collection cost has been an active research topic, with Active Learning (AL) gaining popularity due to its performance and generalisability across various fields. AL is a branch in machine learning that uses an algorithmic approach to identify the most informative instances for human annotation to achieve the desired model performance while minimising annotation costs. A recent study shown that AL can reduce the user labelling effort by up to 46% for personal thermal comfort models [4], demonstrating the potential benefits of AL in this domain.

In this study, we developed a thermal preference-based HVAC control framework enhanced with AL to address the data challenges related to real-world implementations of such OCC systems. The feasibility of our proposed AL-enabled OCC was demonstrated in an EnergyPlus simulation of a real-world testbed, supplemented with the thermal preference data of 58 study occupants. The implications of this work can increase the feasibility of future real-world implementations, enabling OCC and energy-efficient building operations.

## 2. Methodology
### 2.1. Data Collection Setup and Processing

The dataset used in this study was collected from a $50m^2$ testbed constructed in the Building and Construction Authority Academy building in Singapore. The testbed, which represents a single thermal zone, experiences a tropical climate and is conditioned through a variable air volume (VAV) system and two ceiling fans with four levels of speed control. The data was collected over ten consecutive working days involving 58 study participants (29 males, 29 females) between the ages of 21 and 60, and had lived in tropical climates for at least the past three years. Each user is randomly assigned into groups of 5 to 6, where the thermal preference data for each group is collected under different indoor conditions over one working day. Throughout the experiment, indoor air temperature (24°C to 28°C) and air speed (0.1 m/s to 0.8 m/s) were changed once every 30 minutes in a randomised order and indoor conditions were continuously monitored through various sensors. These include indoor air temperature, globe temperature, relative humidity, air speed, total volatile organic compounds, carbon dioxide, and fine particulate matter. The outdoor weather conditions were also measured through a nearby weather station, consisting of outdoor air temperature, outdoor relative humidity, atmospheric pressure, and fine particulate matter. Every 30 minutes, participants were asked to provide feedback about their thermal preferences through a combination of wearables and online surveys. More specifically, a short comfort survey was disseminated to the participants' Apple watches at the $5^{th}$ and $15^{th}$-minute mark, while an online survey was conducted at the $25^{th}$-minute mark to capture the participants' thermal preferences (i.e., Cooler, No Change, Warmer). This resulted in 1,563 valid instances collected. A detailed description of the experiment design is provided in [4].

Based on the thermal comfort dataset collected from the testbed, an Extreme Gradient Boosting (XGB)-based feature selection step was performed to identify the most important features for thermal comfort modelling. This step allows us to reduce the model's likelihood of overfitting and decrease the model's training time. The feature selection approach adopted in this work is based on a previous study [5], which objectively evaluates each feature's usefulness based on its feature ranking and importance score for thermal comfort modelling. Each feature ranking is evaluated using the recursive feature elimination (RFECV) approach, which is a top-down elimination method that recursively eliminates the least relevant features that do not significantly contribute to the model's predictive performance. The approach includes a cross-validation step that splits the dataset into k-folds before obtaining the final rankings for each feature by averaging their rankings across each fold to increase the approach's robustness.

The second step in the framework involved generating the feature importance scores using a tree-based ensemble model. The importance score for each feature was calculated by weighting the decrease in impurity achieved at each attribute split point against the number of samples affected by the split. Finally, by sorting the features based on their feature rankings followed by their feature importance scores, the top five most important features for thermal comfort modelling are indoor temperature, air speed, outdoor temperature, outdoor humidity, and the unique identifier assigned to each participant.

### 2.2. Overview of the Proposed Framework

This study examines two control strategies: 1) AL-enabled OCC with partial preference information and 2) Conventional OCC with complete preference information. The distinction between both strategies lies in the labelling process, where the former strategy uses AL to label the most informative instances while the latter strategy requires all sampled instances to be labelled to generate the individual thermal comfort profiles. Figure 1 presents an overview of the proposed AL-enabled OCC (left) and conventional OCC (right), consisting of six key steps:

 (i) At the beginning of every control time step, sample the thermal preference data of six unique occupants from the thermal comfort dataset based on the current indoor conditions.
 (ii) Apply AL on the sampled thermal preference data and determine which instances should be labelled based on their informativeness towards predicting the occupants' thermal comfort. This step only applies to the AL-enabled OCC strategy.
(iii) Update personal comfort model based on labelled data collected till this control timestep.
(iv) Predict the occupants' thermal preferences across different indoor temperatures and real-time outdoor weather data to generate their thermal comfort profiles.
 (v) Update the zone's optimal setpoint based on the group thermal comfort profile obtained by aggregating the occupants' thermal comfort profiles.
(vi) Run the selected control action in the simulation environment to generate the indoor conditions for the following control timestep.

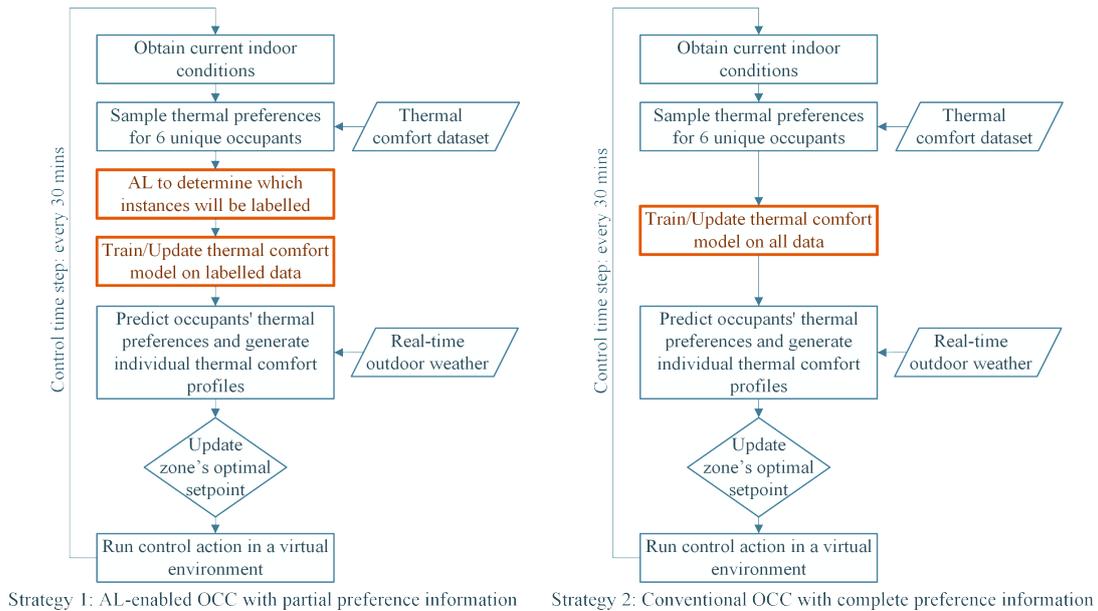

Figure 1: Graphical overview of the proposed OCC strategies.

*2.3. Personal Comfort Model Development with Active Learning*
At each control timestep, AL is applied to the sampled instances from the thermal comfort dataset to identify the most informative instances and query the respective occupants' thermal preferences. The instances are sampled based on their closeness to the current indoor condition to accurately represent the comfort survey responses of the occupants. The AL algorithm selected for this study is based on the Query-by-Committee Sampling (QBC) algorithm, which maintains a committee of model classifiers trained on different subsets of the labelled instances. By voting on the predicted labels of the sampled instances, the instances that resulted in the greatest disagreement among the committee are deemed more informative and selected for human annotation. After the occupants have labelled the informative instances, the newly labelled instances are subsequently used to update the personalised thermal comfort model using a supervised learning approach. The personalised thermal comfort model was developed using an ensemble tree-based model based on the XGB algorithm [6]. The algorithm uses an iterative functional gradient descent method to minimise the loss function in the direction of steepest descent by iteratively introducing a weak classifier in a forward stage-wise fashion to overcome the errors made by previous models [7].

*2.4. Thermal Comfort Profiles*
The resulting model is used to generate the occupants' personalised thermal comfort profiles, which are subsequently aggregated across all occupants to identify the optimal temperature setpoint for the current control timestep. This is achieved by passing into the model information about the occupant's unique ID, the current air speed, outdoor temperature, relative humidity, and a range of indoor temperatures (i.e., 24.5°C to 28°C) to generate the probabilistic distribution of each occupants' preferences (i.e., Cooler, No Change, Warmer) over the range of indoor temperatures. Based on each occupant's thermal comfort profile, their comfort temperatures are determined by the temperature ranges where the probability of feeling comfortable (i.e., No Change) is greater than the probability of preferring a cooler or warmer condition (i.e., Warmer or Cooler). Finally, the optimal setpoint is identified by aggregating the personalised comfort temperatures of all occupants and selecting the highest comfort temperature with the greatest agreement.

*2.5. Virtual Environment Setup*
An EnergyPlus model of the testbed with evidence-based calibration serves as the virtual environment, which has a 7.135% Coefficient of Variance of the Root Mean Square Error (CV(RMSE)) and complies with ASHRAE Guideline 14 [8]. The Singapore's International Weather for Energy Calculation (IWEC) file is also used for annual building energy simulation. At each control timestep, the temperature setpoint of the zone is set to the optimal setpoint determined by the proposed OCC strategies. Subsequently, the indoor conditions are updated to initiate the next control loop. Additionally, the temperature setpoint at the first control timestep was set at 24°C, reflecting Singapore's typical office operating conditions.

## 3. Results and Discussion
*3.1. Comparison of Control Strategies*
Figure 2 presents a comparison of the control actions derived from both AL-enabled OCC and conventional OCC strategies. During the initial week of implementation, notable fluctuations in the optimal temperature setpoints were observed, ranging from 24.5°C to 28°C. This behaviour can be attributed to the limited training data collected in a mild transient condition when training the thermal comfort model at the onset of control implementation. From the third week onwards, the fluctuations in the temperature setpoints decreased substantially, as the thermal comfort profiles reached stability over time, representing steady-state thermal comfort.

Nevertheless, the conventional OCC strategy generally opted for a lower temperature setpoint compared to the AL-enabled OCC strategy. This discrepancy is primarily due to the different amount of thermal preference instances used from the database, despite employing identical methods for developing the thermal comfort profiles. Lastly, around the 37th to 38th day (02-06 to 02-07) of control implementation, the temperature setpoints from both strategies converged to the same value of 27.9°C. This outcome demonstrates that both strategies had, by then, acquired all the necessary thermal comfort information for optimal control. Interestingly, both control strategies stabilized almost on the same date, suggesting that additional thermal comfort information may not necessarily result in faster convergence in control decisions.

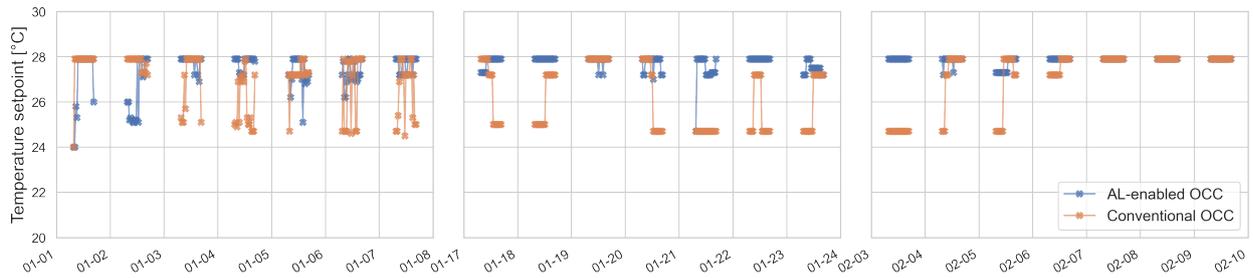

Figure 2: Illustration of the selected control actions for both OCC strategies during the first week (left), third week (middle), and fifth week (right) of the control implementation.

*3.2. Control Performance Evaluation*

The control performance of the proposed AL-enabled OCC is evaluated against conventional OCC by taking into account both building cooling energy reduction and thermal comfort improvement. Moreover, given the availability of elevated air movement provided by ceiling fans, we also included a baseline with a fixed temperature setpoint of 27°C and an air speed of 0.94 m/s. While an air speed of 0.8 m/s is often used as a baseline in accordance with ASHRAE 55 [9], the chosen baseline in this study represents the highest thermal comfort attainable using a rule-based control strategy, as demonstrated in a previous experimental study conducted in the same study area in Singapore [10].

*3.2.1. Cooling Energy Reduction* Figure 3 illustrates the weekly cooling energy consumption for the first eight weeks of control implementation, which consists of district cooling energy, air handling unit (AHU) fan energy, and chilled water pump energy. Given the strong correlation between energy consumption and outdoor weather conditions, the corresponding weekly outdoor temperature and relative humidity are also presented in Figure 3. Both OCC strategies converged around 02-06 to 02-07, resulting in identical cooling energy consumption from that point onwards. According to the annual energy simulation results, the AL-enabled OCC and conventional OCC achieved 4.6% and 3.5% energy reductions over the baseline, respectively. Furthermore, it can be observed from Figure 3 that the AL-enabled OCC consistently reported lower weekly energy consumption compared to conventional OCC during the initial weeks of control implementation prior to convergence due to the selection of higher temperature setpoints.

*3.2.2. Thermal Comfort Acceptability* The thermal comfort acceptability score is calculated at each control timestep for both OCC strategies representing the percentage of occupants who are more likely to be comfortable at the selected temperature setpoint. By averaging the acceptability scores before model convergence, the results for both OCC strategies are similar, with the AL-enabled OCC and conventional OCC reporting an average acceptability score of 98.3% and 99.5%, respectively.

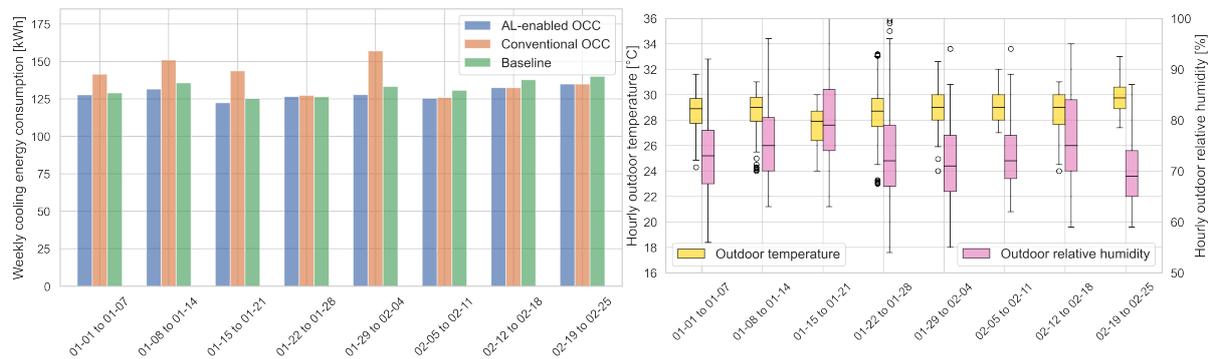

Figure 3: Comparison of weekly cooling energy consumption (left) and weather data (right).

*3.3. Reduction in Labelling Effort*

Lastly, by tracking the model performance of the personal thermal comfort model as more instances are identified and labelled through the AL algorithm, the model was able to match the performance of a fully supervised model with a labelling effort of 69.0%, thereby achieving a 31% reduction in labelling effort.

## 4. Conclusion

In this study, we have demonstrated the feasibility of enhancing thermal preference-based HVAC control with AL to address the data challenges related to real-world implementations of such systems. The results of this study have shown that our AL-enabled OCC strategy significantly reduced the overall labelling effort by 31.0% while achieving a 1.3% increase in energy savings and a high thermal satisfaction score of 98.3% when compared to conventional OCC. Future directions of this work will include the integration of more advanced HVAC control algorithms [11] and personalised air movement preference models to support mixed-mode OCC implementations.

**Acknowledgements**

This study is supported by the National Research Foundation, Singapore, and the Ministry of National Development, Singapore, under its Cities of Tomorrow R&D Programme (CoT Award COT-V4-2020-5) and Johnson Controls (grant number A-8001200-00-00).